\documentclass[conference]{IEEEtran}
\IEEEoverridecommandlockouts
\usepackage{cite}
\usepackage{amsmath,amssymb,amsfonts}
\usepackage{float}
\usepackage{graphicx}
\usepackage{textcomp}
\usepackage[dvipsnames]{xcolor}
\def\BibTeX{{\rm B\kern-.05em{\sc i\kern-.025em b}\kern-.08em
    T\kern-.1667em\lower.7ex\hbox{E}\kern-.125emX}}

    \usepackage{graphicx}
\usepackage{epstopdf}
\usepackage{float}
\usepackage{bm,upgreek}
\usepackage{amsfonts}
\usepackage{enumitem}
\usepackage{mathtools}
\usepackage{multirow}
\usepackage{booktabs}
\usepackage[subnum]{cases}
\usepackage{setspace}
\usepackage{color}
\usepackage{blkarray, bigstrut} 
\usepackage[caption=false,labelformat=simple]{subfig}

\usepackage{xcolor}
\usepackage{pgfplots}
\pgfplotsset{compat=1.5,
  tick label style={font=\small},
  label style={font=\small}}
\usepackage{pgfplotstable}
\usepgfplotslibrary{statistics}
\usepackage{tikz}

\usepackage{pgfplots}
\usepackage{pgfplotstable}
\pgfplotsset{compat=1.7}
\usepgfplotslibrary{groupplots}
\usepackage[toc,acronym]{glossaries}
\usepackage{mathtools,nccmath}
\usepackage[ruled]{algorithm}
\usepackage{algpseudocode}

\algnewcommand{\LineComment}[1]{\State \(\#\) #1}

\begin{document}

\title{Supporting Future Electrical Utilities: Using Deep Learning Methods in EMS and DMS Algorithms\\

\thanks{This work was supported by the Faculty of Technical Sciences in Novi Sad, Department of Power, Electronic and Telecommunication Engineering, within the implementation of the project entitled: "Research aimed at improving the teaching process and development of scientific and professional areas of the Department of Power, Electronic and Telecommunication Engineering. Additionally, this work was supported by the European Union's Horizon 2020 research and innovation programme under Grant Agreement number 856967.}
}

\author{
\IEEEauthorblockN{Ognjen Kundacina}
\IEEEauthorblockA{\textit{The Institute for Artificial Intelligence } \\
\textit{Research and Development of Serbia }\\
Novi Sad, Serbia \\
ognjen.kundacina@ivi.ac.rs}
\and
\IEEEauthorblockN{Gorana Gojic}
\IEEEauthorblockA{\textit{The Institute for Artificial Intelligence } \\
\textit{Research and Development of Serbia }\\
Novi Sad, Serbia \\
gorana.gojic@ivi.ac.rs}
\and
\IEEEauthorblockN{Mile Mitrovic}
\IEEEauthorblockA{\textit{Skolkovo Institute of Science and Technology } \\
Moscow, Russia \\
mile.mitrovic@skoltech.ru}
\and
\IEEEauthorblockN{Dragisa Miskovic}
\IEEEauthorblockA{\textit{The Institute for Artificial Intelligence } \\
\textit{Research and Development of Serbia }\\
Novi Sad, Serbia \\
dragisa.miskovic@ivi.ac.rs}
\and
\IEEEauthorblockN{Dejan Vukobratovic}
\IEEEauthorblockA{\textit{Faculty of Technical Sciences} \\
\textit{University of Novi Sad}\\
Novi Sad, Serbia \\
dejanv@uns.ac.rs}
}

\maketitle

\begin{abstract}
Electrical power systems are increasing in size, complexity, as well as dynamics due to the growing integration of renewable energy resources, which have sporadic power generation. This necessitates the development of near real-time power system algorithms, demanding lower computational complexity regarding the power system size. Considering the growing trend in the collection of historical measurement data and recent advances in the rapidly developing deep learning field, the main goal of this paper is to provide a review of recent deep learning-based power system monitoring and optimization algorithms. Electrical utilities can benefit from this review by re-implementing or enhancing the algorithms traditionally used in energy management systems (EMS) and distribution management systems (DMS).

\end{abstract}


\section{Introduction}
Power systems are undergoing a transition due to the increased integration of renewable energy resources, and as a result they are facing new challenges in their operations. These challenges include the unpredictable nature of renewable energy sources, maintaining stability within the power system, managing the impacts of distributed generation, and the challenges presented by reverse power flows \cite{Aguero2017Challenges}. Consequently, the mathematical formulations of traditional algorithms that solve these problems have become increasingly complex and nonlinear, with larger dimensionality, making their practical implementation and real-time operation more challenging. These algorithms are usually implemented as parts of specialized software solutions, such as energy management systems (EMS) for transmission networks and distribution management systems (DMS) used in distribution networks, which are installed in power system control centres and used by power system operators on a daily basis. Some of the algorithms typically used as EMS and DMS functionalities include state estimation, fault detection and localization, demand and generation forecast, voltage and transient stability assessment, voltage control, optimal power flow, economic dispatch, etc. Increasing amounts of data generated by power systems \cite{Rusitschka2010SGDataCloud} and collected by EMS and DMS are enabling the development of new deep learning-based algorithms to overcome the limitations of traditional ones.

Deep learning is a subfield of artificial intelligence that involves training neural network models to find patterns and make predictions based on the available set of data samples \cite{GoodBengCour16}. Some of the advantages of employing deep learning methods in the field of power systems include:

\begin{itemize}
    \item Speed: Once trained, a deep learning algorithm usually operates quickly, even when processing large amounts of data \cite{Sarker2021DeepLA}. This is crucial for applications where fast decision-making is required, as is the case in many power system operation problems.
    \item Accuracy: Universal approximation theorem \cite{hornik1989UniversApprox} states that neural networks can approximate any function to a desired degree of accuracy, if it consists of a sufficient number of trainable parameters. Practically, this implies that neural networks can be employed to tackle a wide range of problems, including those in power systems, and that different network architectures and sizes can be used to adapt to the complexity of the problem.
    \item Adaptability: Deep learning methods are easily adaptable, meaning that they can be retrained when the underlying data generation process changes \cite{datasetShift}. This makes them suitable for dynamic environments, such as when the power system's operating conditions change.
    \item Robustness: Traditional model-based algorithms can encounter problems when faced with uncertain or unreliable power system parameters \cite{antona2018PSUncertainty}. As a model-free alternative, deep learning methods alleviate these issues by not relying on power system parameters.
    \item Automation: Since deep learning algorithms can learn the responses of human experts in various situations given enough training data, they can be used to reduce the need for human intervention in certain power system tasks. For instance, in applications such as predictive maintenance \cite{Zhang2019PredictiveMainta}, which are integral parts of asset management systems, deep learning can be applied within an automated real-time monitoring system.
\end{itemize}

In the continuation, we shortly introduce the basic deep learning terminology, describe the most common deep learning approaches and review their recent applications in the field of monitoring and optimization of electric power systems.

\section{Deep Learning Fundamentals}
\label{sec:dl}
\begin{figure}[htbp]
    \centerline{\includegraphics[width=3.5in,trim={1.8cm 6cm 7.2cm 3.7cm},clip]{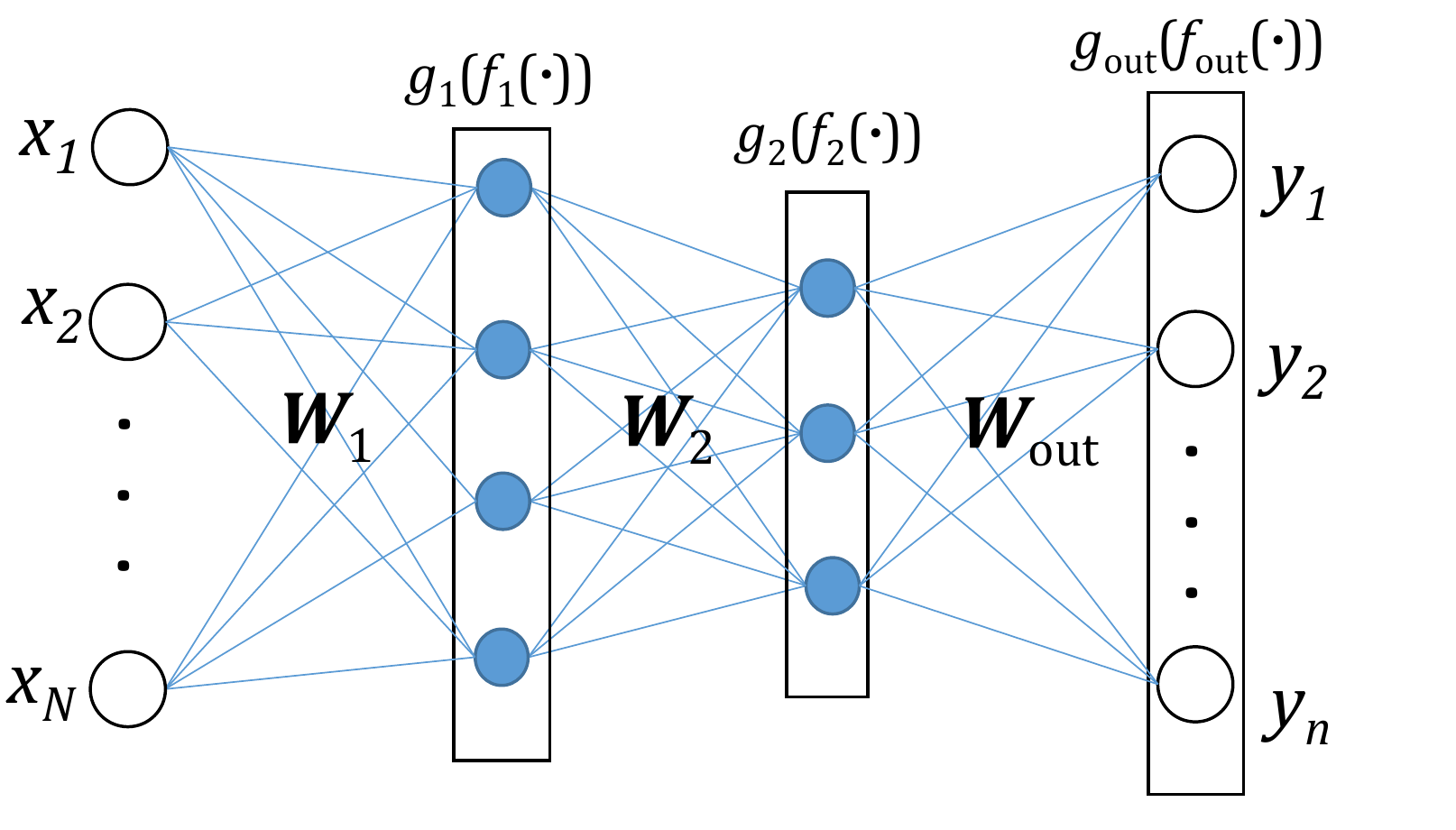}}
    \caption{A simple fully connected neural network containing an input layer, two hidden layers, and an output layer.}
    \label{fig_nn}
\end{figure}

Deep learning is a field of machine learning that involves training neural networks on a large dataset \cite{GoodBengCour16}, with a goal of generating accurate predictions on unseen data samples. Therefore, neural networks can be seen as trainable function approximators, composed of interconnected units called neurons, which process and transmit information. In a simple fully connected neural network, the information processing is organized in layers, where input information from the previous layer is linearly transformed using a function $f_i(\cdot)$, where $i$ denotes the layer index. The linear transformation is defined using a matrix of trainable parameters $\mathbf{W_i}$, i.e., the weights of the connections between the neurons, shown in Fig.~\ref{fig_nn}. Trainable parameters also include biases, which are free terms associated with each neuron, and are omitted in the figure. The information is then passed through a nontrainable nonlinear function $g_i(\cdot)$ to create the outputs of that layer. Inputs and outputs of the whole neural network are denoted as $\mathrm{x}_{j}$ and $y_k$ in Fig.~\ref{fig_nn}, where $j$ and $k$ denote the indices of input and output neurons.

Neural network training assumes adjusting the trainable parameters (i.e., weights and biases of the neurons) using the knowledge in the collected data, so that accurate predictions can be performed based on the new inputs. The training process is formulated as an optimization problem which searches through the trainable parameter space to minimize the distance function between the predicted output and the true output. The problem is usually solved using gradient-based optimization methods such as gradient descent, or some of its variants \cite{KingmaB14Adam}.

In practice, when using deep learning to solve a problem, it is common to train multiple instances with different neural network model structures. This structure is defined by hyperparameters, such as the number of layers and the number of neurons in each layer. By finding the optimal set of hyperparameters, the neural network structure that best fits the problem being solved can be identified. The hyperparameter search can be done manually or with the use of specialized optimization methods \cite{Bergstra2013MakingAS}. Commonly, the collected data is split into three sets: a training set, a validation set, and a test set. The training set is used in a neural network training process, the validation set is used to evaluate the performance of a single training instance, and the test set is used to evaluate the overall performance of the trained model.

Adjusting the deep learning model's architecture to the specific structure of the input data can increase the training speed and performance and reduce the amount of needed training data \cite{Battaglia2018Relational}. This way of exploiting the regularity of the input data space by imposing the structure of the trainable function space is known by the term relational inductive bias \cite{Battaglia2018Relational}. Table~\ref{tblInductiveBias} compares various deep learning models based on their input data structure, the type of neural network layers they use, and the corresponding relational inductive bias.  One of the most successful examples of exploiting relational inductive biases are convolutional neural network (CNN) layers, producing algorithms that surpass human experts in many computer vision tasks. CNNs use the same set of trainable parameters (known as the convolutional kernel) to operate over parts of the input grid data independently, achieving locality and spatial translation invariance. Locality exploits the fact that neighbouring grid elements are more related than further ones, while spatial translation invariance is the ability to map various translations of the input data into the same output. Similarly, recurrent neural networks (RNNs) utilize trainable parameter sharing to process the segments of the sequential data, resulting in a time translation invariant algorithm. The main goal of graph neural networks (GNNs) from the inductive bias perspective is to achieve permutation invariance when applied over graph structured data, so that various matrix representations of the same graph map into the same output. Since ordinary, fully connected neural networks have been widely used for solving power systems problems, we focus on applications of more advanced deep learning architectures.

\begin{table*}[hbpt]
\centering
\caption{Comparison of various deep learning models from the inductive bias perspective.}
\label{tblInductiveBias}
\begin{tabular}{|c|c|c|c|}
\hline
\textbf{Neural network layer type} & \textbf{Input data structure} & \textbf{Relational inductive bias}            & \textbf{Property}                                                                           \\ \hline
Fully connected                                                     & Arbitrary                                                       & Input elements weakly related & -                                                                                  \\ \hline
Convolutional                                                       & Grids, images                                                   & Local relation                                                                  & Spatial translation invariance           \\ \hline
Recurrent                                                           & Sequences                                                     & Sequential relation                                                             & Time translation invariance             \\ \hline
GNN layer                                                           & Graphs                                                          & Arbitrary relation                                                              & Permutation invariance \\ \hline
\end{tabular}
\end{table*}

\section{Convolutional Neural Networks}
Convolutional Neural Networks are a well studied class of deep learning algorithms, primarily designed for analysing spatial patterns in grid-structured data such as images \cite{GoodBengCour16}. They consist of multiple convolutional layers, each of which acts as a trainable convolutional filter that extracts local information from the image, transforms it into more abstract, grid-shaped representations, and feeds it into the succeeding layer. Applying multiple CNN layers enables CNN to extract useful features from an image, which can then be used for various tasks such as classification or regression.

Although power system data is not inherently arranged in the format of an image, CNNs have been effectively used to address power system problems, mostly involved with processing data sequences. To meet the requirements of CNNs, power system data is transformed and reshaped in various ways, some of which include:
\begin{itemize}
    \item One approach for dealing with the time-varying nature of power systems is to utilize 1D CNNs on univariate time series data. For example, in study \cite{Poudyal2021ConvolutionalNN}, 1D CNNs were used to predict power system inertia using only frequency measurements. The process involves stacking time series of changes in frequency measurements, along with their rates of change, into a one-dimensional array and then processing it using 1D CNNs.
    \item A more effective method is to group signals into a matrix, where each row represents a single univariate signal. By using a 2D CNN to process this matrix, we can perform multivariate time series analysis, which allows us to analyse patterns across multiple time series and how they interact with each other. This approach has been used in recent research, such as in the study \cite{Alqudah2022CnnFaults}, to detect faults in power systems through analysing series of voltage, current, and frequency measurements.
    \item Time series data can be subjected to time-frequency transformation, allowing for analysis of the frequency content of the signal while maintaining its temporal localization. These transformations can be visually represented in two dimensions, and therefore can be analysed using various image processing tools, including CNNs. For instance, in \cite{Scalograms2020} a CNN was trained to classify faults in power systems by analysing 2D scalograms, which were generated by applying the continuous wavelet transform to time series of phasor measurements.
    \item Another approach is to use a CNN over the matrix of electrical quantities created for a single time instance, where each row contains the values of a specific electrical quantity for each power system element. This approach, which does not consider time series data, has been shown to be effective in certain applications. The study \cite{DCOPF_CNN} solves the DC optimal power flow problem by using this approach and taking node-level active and reactive power injections as inputs, with labels obtained using the traditional DC optimal power flow approach.
\end{itemize}

It's important to note that these approaches use only aggregated inputs from all the elements of the power system, without considering the connectivity between them.

\section{Recurrent Neural Networks}
Recurrent neural networks represent a significant development in deep learning algorithms, particularly in the processing of sequential data such as speech, text, and time series. \cite{GoodBengCour16}. Each of the recurrent layers acts as a memory cell that takes in information from previous steps in the sequence, processes it, and generates a hidden state representation that is passed on to the next step. The final hidden state of RNNs encapsulates the information of the entire input sequence and can be applied to tasks such as natural language processing, speech recognition, and time-series prediction. While 1D CNNs are limited to fixed length sequences, meaning that all time series in the training and test samples must have the same number of elements, RNNs are adaptable to varying sequence lengths, making them more versatile and useful for analysing sequential data.

The fundamental building blocks of RNNs are memory units, such as gated recurrent units (GRUs) and long short-term memory units (LSTMs) \cite{LSTMvsGRU}. These architectures are created to tackle the challenge of longer-term dependencies in sequential data. Both GRUs and LSTMs include an internal memory, which allows them to selectively retain or discard information from previous steps in the sequence, thus enhancing their ability to handle inputs of varying lengths. LSTMs are more complex and powerful, capable of handling longer-term dependencies, while GRUs are computationally simpler and faster, yet may not be as effective in certain tasks.

In the field of power demand and generation forecasting, various time series prediction algorithms, including RNNs, have been utilized. One recent study, \cite{MissingDataTolerantPVForecasting} uses LSTM RNNs to predict multistep-ahead solar generation based on recorded measurement history while also addressing missing records in the input time series. RNNs can also be used to predict the flexibility of large consumers' power demand in response to dynamic market price changes, as demonstrated in \cite{SiameseLSTMDemandFlexibility}. This approach combines two LSTM RNNs, one for predicting market price and the other for predicting a consumer's demand flexibility metric, with a focus on uncommon events such as price spikes. An interesting technical aspect of this method is that the two RNNs share some LSTM-based layers, resulting in more efficient and faster training, as well as improved prediction capabilities.

RNNs can also be applied to other data available in DMS and EMS, unrelated to power and energy. The work \cite{BidirectionalLSTMVoltageStability} proposes using an RNN to classify the voltage stability of a microgrid after a fault, using time series of measurement deviations, providing power system operators with valuable information, needed to take corrective actions. The employed RNN architecture is the bidirectional LSTM, which processes the time series data in both forward and backward directions, which allows the RNN to consider both past and future context in each step of the sequence when making predictions. In the study \cite{FELLNER2022100851}, the authors evaluate different deep learning models for detecting misconfigurations in power systems using time series of operational data. They compare GRU RNN, LSTM RNN, the transformer architecture \cite{attentionAllYouNeed}, which has been successful in natural language processing tasks, and a hybrid RNN-enhanced transformer \cite{RTransformer}. They find that the RNN-enhanced transformer is the most effective architecture, highlighting the potential of attention-based architectures for solving time series problems in power systems.

\section{Graph Neural Networks}
Graph Neural Networks, particularly spatial GNNs that utilize message passing, are an increasingly popular deep learning technique that excels at handling graph structured data, which makes them particularly well-suited for addressing a wide range of power systems problems. Spatial GNNs process graph structured data by repeatedly applying a process called message passing between the connected nodes in the graph \cite{GraphRepresentationLearningBook}. The goal of GNNs is to represent the information from each node and its connections in a higher-dimensional space, creating a vector representation of each node, also known as node embeddings. GNNs are made up of multiple layers, each representing one iteration of message passing. Each message passing iteration is performed by applying multiple trainable functions, implemented as neural networks, such as a message function, an aggregation function, and an update function. The message function calculates the messages being passed between two node embeddings, the aggregation function combines the incoming messages in a specific way to create an aggregated message, and the update function calculates the update to each node's embedding. This process is repeated a predefined number of times, and the final node embeddings are passed through additional neural network layers to generate predictions.

GNNs have several advantages over the other deep learning methods when used in power systems. One of them is their permutation invariance property, which means that they produce the same output for different representations of the same graph by design. GNNs are able to handle dynamic changes in the topology of power systems and can effectively operate over graphs with varying numbers of nodes and edges. This makes them well suited for real-world power systems, which may have varying topologies. Additionally, GNNs are computationally and memory efficient, requiring fewer trainable parameters and less storage space than traditional deep learning methods applied to graph-structured data, which is beneficial in power system problems where near real-time performance is critical. Spatial GNNs have the ability to perform distributed inference with only local measurements, which makes it possible to use the 5G network communication infrastructure and edge computing to implement this effectively \cite{kundacina5G2022}. This enables real-time and low-latency decision-making in large networks as the computations are done at the network edge, near the data source, minimizing the amount of data sent over the network.

GNNs have recently been applied to a variety of regression or classification tasks in the field of power systems. The work \cite{chen2020FaultLocation} proposes using GNNs over the bus-branch model of power distribution systems, with phasor measurement data as inputs, to perform the fault location task by identifying the node in the graph where the fault occurred. The use of GNNs for assessing power system stability has been explored in \cite{ZHANG2022Stability}, where the problem is formulated as a graph-level classification task to distinguish between rotor angle instability, voltage instability, and stability states, also based on power system topology and measurements. The paper \cite{ARASTEHFAR2022LoadFOrecasting} presents a hybrid neural network architecture which combines GNNs and RNNs to address the Short-Term Load Forecasting problem. The RNNs are used to process historical load data and provide inputs to GNNs, which are then used to extract the spatial information from users with similar consumption patterns, thus providing a more comprehensive approach to forecast the power consumption. In \cite{Zhao2022Transient} the authors propose a GNN approach for predicting the power system dynamics represented as time series of power system states after a disturbance or failure occurs. The GNN is fed with real-time measurements from phasor measurement units that are distributed along the nodes of the graph. In \cite{takiddin2022FalseAttacks} GNNs are applied over varying power system topologies to detect unseen false data injection attacks in smart grids.

In the previously mentioned studies, GNNs have been applied to the traditional bus-branch model of power systems, however, a recent trend in the field has been to apply GNNs over other topologies representing the connectivity in power system data. For example, GNNs have been used in combination with heterogeneous power system factor graphs to solve the state estimation problem, both linear \cite{kundacina2022state} and nonlinear \cite{kundacina2022NonlinearSE}. In these approaches, measurements are represented using factor nodes, while variable nodes are used to predict state variables and calculate training loss. These approaches are more flexible regarding the input measurement data compared to traditional deep learning-based state estimation methods because they provide the ability to easily integrate or exclude various types of measurements on power system buses and branches, through the addition or removal of the corresponding nodes in the factor graph. A different approach that does not use the GNN over the traditional bus-branch model is presented in \cite{Yuan2022LearningInterractions}. The proposed method solves the power system event classification problem based on the collected data from phasor measurement units. The approach starts by using a GNN encoder to infer the relationships between the measurements, and then employs a GNN decoder on the learned interaction graph to classify the power system events.

\section{Deep Reinforcement Learning}
\begin{figure}[htbp]
    \centerline{\includegraphics[width=2.6in,trim={2cm 2.0cm 2.2cm 2.8cm},clip]{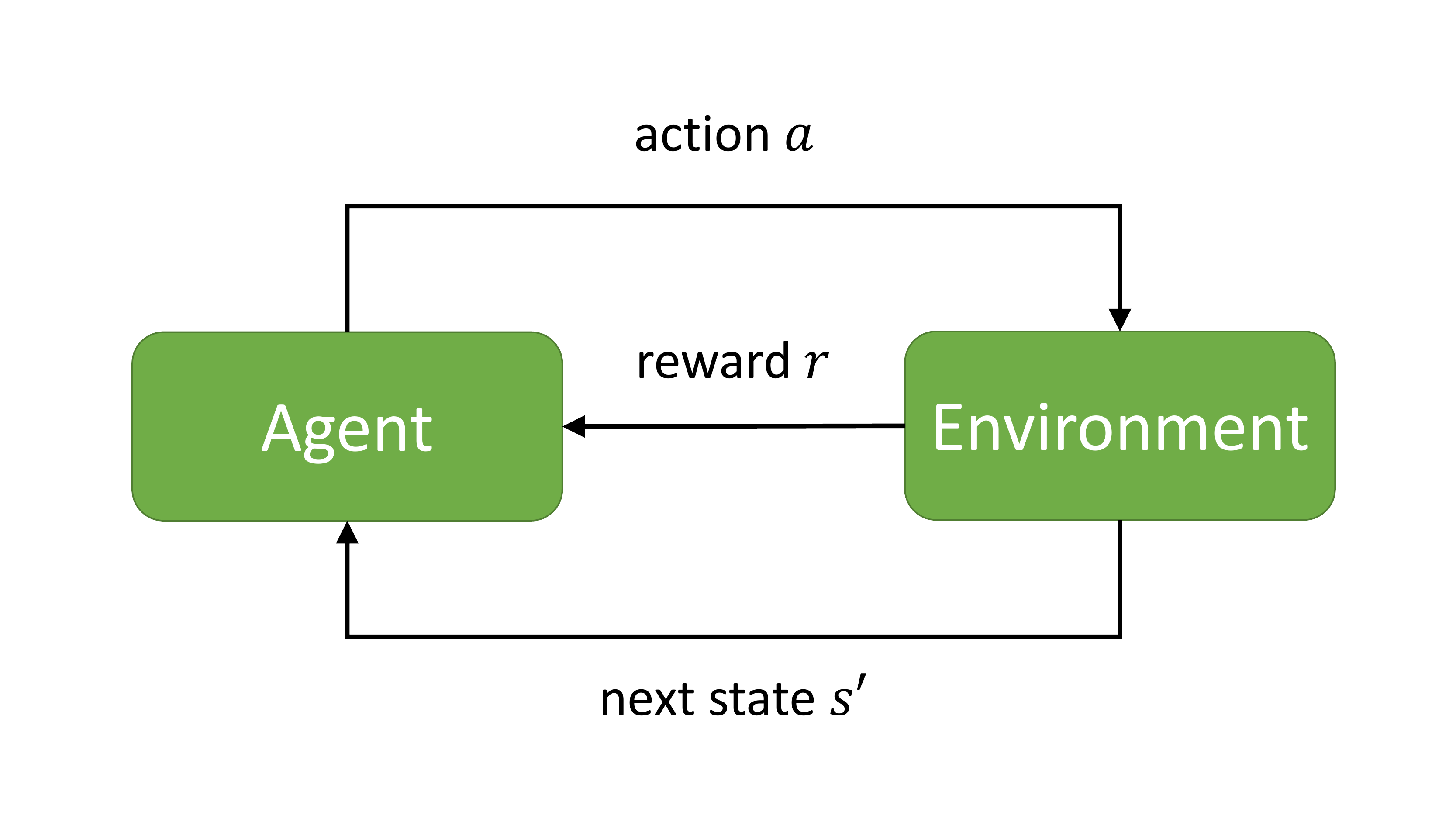}}
    \caption{The agent-environment interaction process.}
    \label{agentEnvInterraction}
\end{figure}

So far, we have reviewed deep learning methods that are inherently suited for predicting discrete or continuous variables based on a set of inputs. In contrast, deep reinforcement learning (DRL) methods have a direct goal of long-term optimization of a series of actions that are followed by immediate feedback\cite{Sutton1998}. Therefore, DRL methods are powerful tools for multi-objective sequential decision-making, suitable for application in various EMS and DMS functionalities that involve power system optimization \cite{Glavic2019}. In the DRL framework, the agent interacts with the stochastic environment in discrete time steps and the goal is to find the optimal policy that maximizes the long-term reward while receiving feedback about its immediate performance. The agent receives state variables from the environment, takes an action, receives an immediate reward signal and the state variables for the next time step, as shown in Fig.~\ref{agentEnvInterraction}. The DRL training process involves many episodes that include agent-environment interaction, during which the agent learns by trial and error. Using the collected data from these episodes, the agent is able to predict the long term rewards in various situations using neural networks, and these predictions are then used to generate an optimal decision-making strategy.

There are many studies that apply DRL in the field of power system optimization and control. Some of the examples include distribution network reconfiguration for active power loss reduction \cite{kundavcina2022solving}, Volt-VAR control in electrical distribution systems \cite{wang2021SafeVVC}, frequency control in low-inertia power systems \cite{stanojev2021FFC}, and so on. In these studies, an RL agent receives various electrical measurements as state information and takes a single multidimensional action per time step, which includes both discrete and continuous set points on controllable devices within a power system.

A recent trend in the power system research is transitioning from single agent to multi-agent deep reinforcement learning (MADRL), which is based on coordinating multiple agents operating together in a single environment using the mathematical apparatus developed in the field of game theory \cite{MultiAgentRL}. MADRL relies on centralized training and decentralized execution concept, where a centralized algorithm is responsible for training all the agents at once, allowing for coordination and cooperation among the agents. This centralized training approach results in faster real-life execution due to significantly reduced communication delays during decentralized execution, where each agent can act independently based on the knowledge acquired during the centralized training. Reducing these communication delays is particularly important in large transmission power systems where the individual agents may be significantly geographically separated.

For example, a decentralized Volt-VAR control algorithm for power distribution systems based on MADRL is proposed in \cite{MARLVVC}. In this algorithm, the power system is divided into multiple independent control areas, each of which is controlled by a corresponding DRL agent. These agents observe only the local measurements of electrical quantities within their corresponding area, and the action of each agent contains set points on all the reactive power resources in that area. Similarly, in \cite{PowerNet}, a MADRL algorithm is used to solve the secondary voltage control problem in isolated microgrids in a decentralized fashion by coordinating multiple agents, each of which corresponds to a distributed generator equipped with a voltage-controlled voltage source inverter. The action of each agent is a single secondary voltage control set point of the corresponding generator. The fundamental difference compared to \cite{MARLVVC} is that the agent in \cite{PowerNet} uses not only the local measurements of electrical quantities for the state information, but also messages from the neighbouring agents, leading to improved performance. Work \cite{MARL_EconomicDispatch} proposes using a MADRL algorithm to perform the economic dispatch, which minimizes the overall cost of generation while satisfying the power demand. The agent models an individual power plant in a power system, with the action being the active power production set point. Another example of using MADRL for an economic problem in coupled power and transportation networks is given in \cite{MADRL_EVCharging}. A MADRL method is proposed to model the pricing game and determine the optimal charging pricing strategies of multiple electric vehicle charging stations, where each individually-owned EV charging station competes using price signals to maximize their respective payoffs. In all the aforementioned works, multiple agents are trained in a centralized manner to optimize the reward function defined globally based on the nature of the particular problem at hand.

\section{Conclusions}
Deep learning has demonstrated great potential to improve various aspects of both EMS and DMS, including power system monitoring tasks such as stability assessment, state estimation and fault detection, as well as for power system optimization tasks like Volt-Var optimization, distribution network reconfiguration, etc. Reviewed studies indicate that these methods exhibit high levels of accuracy and improved performance when compared to traditionally used techniques. One of the current trends in the field is the use of graph neural networks and multi-agent deep reinforcement learning. As the field continues to evolve, it is expected that more research and development will be conducted in these areas, with a focus on implementing these techniques in real-world power systems to demonstrate their practical potential.

\bibliographystyle{IEEEtran}
\bibliography{cite}

\end{document}